\documentclass{article}

\usepackage{arxiv}

\usepackage[utf8]{inputenc} % allow utf-8 input
\usepackage[T1]{fontenc}    % use 8-bit T1 fonts
\usepackage{hyperref}       % hyperlinks
\usepackage{url}            % simple URL typesetting
\usepackage{booktabs}       % professional-quality tables
\usepackage{amsfonts}       % blackboard math symbols
\usepackage{nicefrac}       % compact symbols for 1/2, etc.
\usepackage{microtype}      % microtypography
\usepackage{lipsum}		% Can be removed after putting your text content
\usepackage{graphicx}
\usepackage{natbib}
\usepackage{doi}
\usepackage{enumitem}

\title{The semantic landscape paradigm for neural networks}

\author{ \href{https://orcid.org/0000-0002-0706-5320}{\includegraphics[scale=0.06]{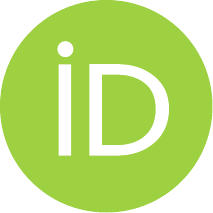}\hspace{1mm}Shreyas Gokhale} \\
	Department of Physics\\
	Massachusetts Institute of Technology\\
	Cambridge, MA 02139 \\
	\texttt{gokhales@mit.edu} \\
}

\renewcommand{\headeright}{}
\renewcommand{\undertitle}{}
\renewcommand{\shorttitle}{}

\begin{document}
\maketitle

\begin{abstract}
    Deep neural networks exhibit a fascinating spectrum of phenomena ranging from predictable scaling laws to the unpredictable emergence of new capabilities as a function of training time, dataset size and network size. Analysis of these phenomena has revealed the existence of concepts and algorithms encoded within the learned representations of these networks. While significant strides have been made in explaining observed phenomena separately, a unified framework for understanding, dissecting, and predicting the performance of neural networks is lacking. Here, we introduce the semantic landscape paradigm, a conceptual and mathematical framework that describes the training dynamics of neural networks as trajectories on a graph whose nodes correspond to emergent algorithms that are instrinsic to the learned representations of the networks. This abstraction enables us to describe a wide range of neural network phenomena in terms of well studied problems in statistical physics. Specifically, we show that grokking and emergence with scale are associated with percolation phenomena, and neural scaling laws are explainable in terms of the statistics of random walks on graphs. Finally, we discuss how the semantic landscape paradigm complements existing theoretical and practical approaches aimed at understanding and interpreting deep neural networks.    
\end{abstract}

% keywords can be removed
\keywords{Artificial Intelligence \and Machine Learning \and Statistical Physics}

\section{Introduction}
We live in an age in which the readers of this manuscript can never quite be sure whether it was written by a human or a large language model\footnote{For the record, this manuscript has been written in its entirety by a human named Shreyas Gokhale.}. This sobering fact at once highlights the stunning technological advances in artificial intelligence (AI) in recent years, as well as the urgent need to understand the inner workings of the deep neural networks that made these advances possible. Understanding neural networks requires us to know \textit{what} the networks are capable of doing, \textit{how} their performance depends on various factors, and finally \textit{why} they succeed or fail in different circumstances. Over the last few years, researchers have begun to answer these questions through an impressive set of experimental results as well as theoretical ideas. Several studies have reported empirical observations of power law scaling of the test loss as a function of network size (number of parameters), dataset size (number of training examples), compute budget as well as training time (\cite{hestness2017deep,rosenfeld2019constructive,kaplan2020scaling,gordon2021data,zhai2022scaling,hoffmann2022training}). Subsequently, a number of distinct theoretical proposals have been put forward to explain the observed power laws (\cite{bahri2021explaining,hutter2021learning,sharma2022scaling,maloney2022solvable,michaud2023quantization}). 

In addition to these ubiquitous power laws, which provide a precise account of the improvement of model performance with scale, a number of studies have also reported the sudden emergence of new capabilities in neural networks. New capabilities can emerge either as a function of time, a phenomenon known as grokking (\cite{power2022grokking,thilak2022slingshot,liu2022omnigrok}), or as a function of the network size, which we will refer to in this paper as `emergence with scale' (\cite{wei2022emergent,ganguli2022predictability}). Despite their prevalence, the nature of emergent phenomena in deep neural networks remains far from understood. While some works support the view that grokking is associated with a phase transition (\cite{vzunkovivc2022grokking,liu2022towards}), others have contested that sudden emergence is only an illusion stemming from the metric used for quantifying loss, and that the network's performance is in fact evolving gradually (\cite{schaeffer2023emergent}). Furthermore, networks can make hidden progress before the clear manifestation of emergence, that remains undetected in loss curves (\cite{barak2022hidden}). 

A promising approach towards disentangling these seemingly contradictory observations about the emergence is to look under the hood of the network, and determine what the networks in question are actually learning to do. This line of inquiry has led to some truly remarkable discoveries on the nature of learned representations in neural networks. Using network dissection (\cite{bau2017network}), \cite{bau2020understanding}) showed that individual neurons in convolutional neural networks (CNNs) can represent concepts associated with physical objects. By reverse engineering neural networks (\cite{cammarata2020thread:}), a number of studies have shown that transformer language models can learn a variety of algorithms for tasks including text prediction (\cite{elhage2021mathematical,olsson2022context}), modular addition (\cite{nanda2023progress,zhong2023clock}), and group operations (\cite{chughtai2023toy}). Collectively, these findings establish the fact that deep neural networks are capable of learning well-defined computations that confer enhanced prediction capabilities. Based on this idea, (\cite{michaud2023quantization}) have recently proposed the quantization model to explain scaling as well as emergence in language models. 

Owing to the preponderance of performance metric data and the growing literature on mechanistic insights, several independent theoretical approaches aimed at explaining various aspects of neural network phenomenology have been proposed, as discussed in preceding paragraphs. While all of these approaches are immensely valuable, they follow from distinct conceptual frameworks. This makes it difficult to understand how different approaches are related to each other, and if and when their domains of applicability intersect. A unified conceptual framework that integrates quantitative neural network performance metrics with the qualitative demands of interpretability is highly desirable, as it will provide a common language for theoretical discourse, and direct further developments in the field. 

In the present work, we introduce such a unified conceptual framework by defining a semantic landscape in the space spanned by internal algorithms learned by neural networks, which we term heuristic models. We then proceed to demonstrate how several distinct qualitative and quantitative aspects of neural network phenomenology find a natural and consistent description in terms of trajectories on the semantic landscape. The rest of the paper is organized as follows: In Section $2$, we provide mathematical definitions for key concepts associated with the semantic landscape. In Section $3$, we illustrate how grokking and emergence with scale can be described in terms of first passage and percolation phenomena on the semantic landscape. In Section $4$, we derive scaling laws for test loss as a function of network size, dataset size, and training time, for a particular choice of semantic landscape topography. Finally, in Section $5$, we discuss how the semantic landscape paradigm connects to existing approaches, how it can be developed further, and what its implications are for future research.

\section{The semantic landscape paradigm}
\subsection{From the loss landscape to the semantic landscape}
At its core, neural network learning is an optimization problem with many interacting degrees of freedom. This fact has long attracted the attention of statistical physicists, as it enables them to apply their formidable repertoire of theoretical tools to a challenging problem of tremendous practical importance \cite{mehta2019high}. In particular, the concept of a loss landscape defined by the network loss as a function of neural weights plays a central role in our understanding of neural networks, owing to its similarities with well-studied objects such as energy landscapes in physics (\cite{debenedetti2001supercooled}) and chemistry (\cite{onuchic1997theory}), and fitness landscapes in evolutionary biology (\cite{de2014empirical}). In particular, concepts such as scaling and the renormalization group (\cite{geiger2020scaling,bahri2021explaining,roberts2022principles}),  glasses (\cite{baity2018comparing}), and jamming (\cite{spigler2019jamming}), have provided significant insights into the nature of training dynamics. The loss landscape is immensely useful, as it enables us to understand learning dynamics by analogy with well-understood physical phenomena that we have much better intuition for. However, recent studies on mechanistic interpretability (\cite{nanda2022mechanistic,zhong2023clock}), have brought to light a significant gap in our understanding of how neural networks function. Specifically, we do not yet understand how dynamics on the loss landscape can produce outputs that look as if they follow from a sequence of interpretable logical steps. To bridge this gap, we need to develop a coarse-grained description that can on one hand retain connections with loss landscapes, and on the other hand, take direct cognizance of the emergent algorithms that are being discovered in a growing number of studies. The purpose of this paper is to introduce such a description in terms of the semantic landscape, and demonstrate that it can consistently explain a number of disparate neural network phenomena within the same conceptual framework. 

\subsection{Heuristic models}
Understanding how neural networks function is synonymous with understanding how information processing occurs between their input and output layers. Mechanistic interpretability studies strongly suggest that this information processing is analogous to the execution of emergent algorithms encoded within the learned representations of neural networks. The semantic landscape paradigm formalizes this notion with the help of the following conjecture

\begin{itemize}
    \item[] For every neural network, there exists at least one function that maps subsets of the set of network parameters to well formed formulas within a formal language.   
\end{itemize}

A formal language $L$ is defined as a set of words formed using symbols from an alphabet $\Sigma$, according to a given set of rules. A well-formed formula $f$ within $L$ is defined as a finite sequence of symbols belonging to $\Sigma$, that has been constructed using the rules of $L$. Thus, the formal language $L$ can be identified with the set of all well-formed formulas in $L$. We also note that the `formal language' discussed here is associated with internal representations of the neural network, and is not related in any way to the programming language in which the neural network code is written. Mathematically, the conjecture above can be expressed as follows: 

\begin{itemize}
    \item[] \textbf{C1}: Given a neural network with a set of parameters $W = \{w_i|i\in \mathbb{N}^{+},i\leq N\}$, and a formal language $L$ containing the set of well formed formulas $\mathcal{S}_F = \{f_i |i\in \mathbb{N}^{+},i\leq N_L\}$, $\exists \mathcal{G} \neq \emptyset$, such that $\forall G_i \in \mathcal{G}, \\G_i:W^{m}\mapsto\mathcal{S}_F$ for some $m\in\mathbb{N}^{+}$. 
\end{itemize}

Why would such functions $G_i$ that map neural weights to well-formed formulas be of interest? The reason is that it is always possible to uniquely encode well-formed formulas in any language as numbers using schemes such as G\"odel numbering (\cite{godel1931formal}). The functions $G_i$ would therefore ensure a mathematical equivalence between the state of the neural network as described by a set of neural weights, and the state of the neural network as described by a set of well-formed formulas in a formal language. Since algorithms are by definition, sets of well-formed formulas within a given formal language, \textbf{C1} implies that the collective neural activity within all the hidden layers for a fixed set of neural weights is mathematically equivalent to the execution of an algorithm in some formal language $L$.

The conjecture \textbf{C1} is strongly supported by empirical evidence across various models and datasets in the form of concept neurons (\cite{bau2020understanding}), induction heads (\cite{elhage2021mathematical}), learned algorithms for mathematical operations (\cite{nanda2023progress,chughtai2023toy,zhong2023clock}), and auto-discovered quanta (\cite{michaud2023quantization}). Indeed, these studies have also shown that beyond the formulation of logical propositions, neural networks are capable of integrating these propositions into emergent heuristic algorithms that enable the network to generate correct outputs for a majority of inputs. We use the term `heuristic' to emphasize the fact that the emergence of a logical procedure for determining the output need not imply an innate knowledge or understanding of underlying mathematical or factual truths on the network's part. Within the semantic landscape paradigm, we formalize the notion of emergence of such internal heuristic algorithms by defining the concept of a heuristic model as follows:

\begin{itemize}
    \item[] \textbf{Def. 1}: A heuristic model $\mathcal{M}_H$ is an algorithm in $L$ composed of a finite sequence of well formed formulas $f_i \in \mathcal{S}_F$, that generates a unique conditional probability distribution $\mathcal{P}_{H}(y|x)$ over all possible outputs $y$, for every possible input $x$.
\end{itemize}

\begin{figure}
	\centering
    \includegraphics[width= \textwidth]{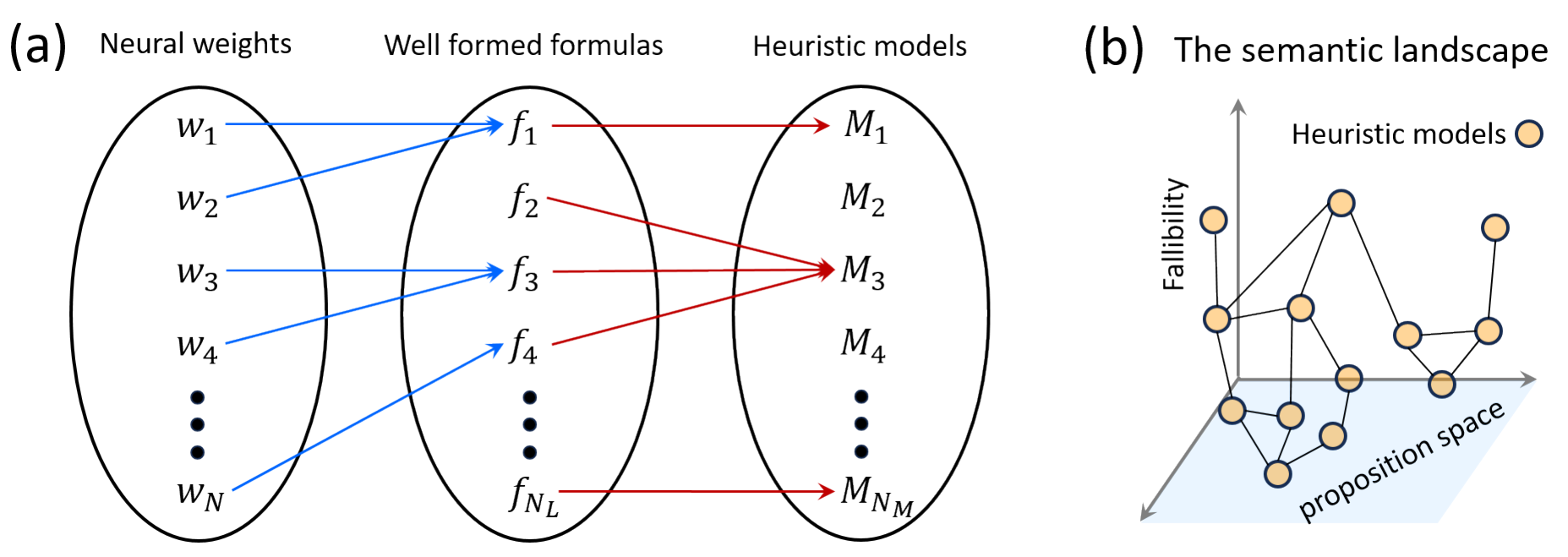}
	\caption{\textbf{Constructing the semantic landscape:}(a) Schematic illustration of the coarse-graining from neural weights to well formed formulas, and their subsequent combination into heuristic models, as proposed in conjecture \textbf{C1}. (b) Schematic of the semantic landscape defined by the variation of fallibility $\mathcal{F}$ over heuristic models in proposition space. At a specific point in training, neural weights attain specific values, and the network occupies a single node on the graph of heuristic models. At a later time, the neural weights change, enabling the network to occupy a different node on the graph of heuristic models.}
	\label{fig1}
\end{figure}

We refer to output distributions rather than unique outputs in \textbf{Def 1}, as for some tasks such as natural language prediction, there may not be a unique `correct' answer, and the range of appropriate responses will typically derive from some distribution over the alphabet of that natural language. The hierarchy of steps from weights to well formed formulas to heuristic models detailed above, is shown schematically in Fig. \ref{fig1}a. Several comments about the notion of heuristic models are in order. First, while heuristic models are collections of well formed formulas in $L$, this does not necessarily imply that they will make correct predictions for inputs, or result in low test loss. In fact, in general, most heuristic models are likely to be \textit{bad} algorithms that result in \textit{high} test loss. As training progresses, the network keeps improving its performance by discovering increasingly better heuristic models. A successfully trained network is one that has ``found'' a heuristic model that is capable of generalizing over a large fraction of possible test data. For instance, in the example of modular addition studied by \cite{nanda2023progress}, the network initially uses a bad heuristic model that only memorizes training data, but later finds a clever mathematical algorithm that can generalize well across all inputs. 

Another important point to note is that a network of a fixed size will in general not be able to formulate all possible heuristic models. This follows directly from \textbf{C1}, because as we increase the number of parameters $N$, the number of ways in which weights can be combined into well formed formulas also increases. In colloquial terms, increasing the network size increases the ``vocabulary'' of the network. We will see in Section $3$ that this increased capacity for formulating algorithms with network size is crucial for the emergence of new capabilities with scale.  

\subsection{Fallibility and the semantic landscape}
Effectively, the foregoing discussion implies that the training dynamics of a neural network can be viewed as dynamics on a graph whose vertices are heuristic models. At a given point in training, the network occupies a single node on the graph of heuristic models. As training progresses and neural weights are adjusted, the network can adopt a different heuristic model, which corresponds to a transition from one node to another. The transition rates to move from one heuristic model to another are governed by a number of factors including the network size and dataset size, data quality, and hyperparameters associated with the network architecture and training algorithm. In principle, one can adopt a bottom-up approach to compute these transition rates by systematically coarse-graining over the network parameters, for specific choices of network architecture and training algorithm. At present, however, such a task is prohibitively difficult to accomplish. Instead, in this introductory version of the semantic landscape paradigm, we implement a top-down approach by making simplified choices for transition rates that are motivated by, and consistent with, network performance measurements. In order to make such choices, we need to quantify precisely how \textit{good} or \textit{bad} a given heuristic model is. We do so by defining the fallibility $\mathcal{F}$ of a given heuristic model $\mathcal{M}_H$ as

\begin{equation} \label{eq:1}
    \mathcal{F}(\mathcal{M}_H) = \sum_{x\in\mathcal{I}} D_{KL} \big(\mathcal{P}_{GT}(y|x) \parallel \mathcal{P}_H(y|x)\big)
\end{equation}

where $\mathcal{P}_{GT}(y|x)$ is the ground truth probability distribution over all possible outputs $y$ for input $x \in \mathcal{I}$, and $D_{KL}(P \parallel Q)$ is the Kullback-Leibler divergence of $P$ from $Q$. From this definition, it is clear that $\mathcal{F}(\mathcal{M}_H)\geq 0 \hspace{1mm} \forall \mathcal{M}_H$, and $\mathcal{F}(\mathcal{M}_H)=0$ if and only if the output distribution generated by $\mathcal{M}_H$ is indistinguishable from the ground truth for every $x\in \mathcal{I}$. Thus, the lower the value of $\mathcal{F}$, the better the heuristic model is. It is important to note that the fallibility of a model is defined over \textit{all possible} inputs, and not just those in the training and test datasets. It is therefore distinct from the measured loss. However, if the training data quality is good, measured test loss will be lower for models of low fallibility compared to those of high fallibility. This can be seen explicitly by considering the set of all possible inputs $\mathcal{I}$ as a union of the training dataset $\mathcal{I}_{\rm{train}}$, test dataset $\mathcal{I}_{\rm{test}}$ and the remaining unseen data $\mathcal{I}_{\rm{unseen}}$. Here, $\mathcal{I}_{\rm{unseen}}$ refers to the set of inputs that the neural network has not seen either during training or during testing. 

\begin{equation}\label{eq:2}
    \mathcal{F} = \sum_{x\in\mathcal{I}_{\rm{train}}} \sum_{x\in\mathcal{I}_{\rm{test}}} \sum_{x\in\mathcal{I}_{\rm{unseen}}} D_{KL} \big(\mathcal{P}_{GT}(y|x) \parallel \mathcal{P}_H(y|x)\big) = \mathcal{F}_{\rm{train}} + \mathcal{F}_{\rm{test}} + \mathcal{F}_{\rm{unseen}}
\end{equation}

Thus, if the network has simply memorized the training dataset, it will have low $\mathcal{F}_{\rm{train}}$, and hence, low measured training loss, but will generally have high $\mathcal{F}_{\rm{test}}$, and therefore high measured test loss. By contrast, a network that has successfully learned to generalize over all inputs will yield low values for $\mathcal{F}_{\rm{train}}$, $\mathcal{F}_{\rm{test}}$, as well as  $\mathcal{F}_{\rm{unseen}}$, and will therefore have low test loss. Defined in this manner, fallibility of a heuristic model is analogous to the energy of an excited state, measured with respect to the energy of the ground state. In analogy with energy landscapes defined over the space of configurations for glasses (\cite{debenedetti2001supercooled}) and proteins (\cite{onuchic1997theory}), or fitness landscapes defined over genomes (\cite{de2014empirical}), we can define the semantic landscape of neural networks in terms of the variation of fallibility $\mathcal{F}$ in the space of well formed formulas $f_i \in \mathcal{S}_F$, such that each ``configuration'' $\{f_i\}$ corresponds to a particular heuristic model $\mathcal{M}_{H}$ (Fig. \ref{fig1}b). An immediate corollary of this definition is that the neural network can potentially get trapped in local minima of the semantic landscape, which prevent the network from improving its performance. This situation implies that small changes to the heuristic model adopted by the network will necessarily lead to worse predictions, and therefore higher loss. In order to make accurate predictions over a majority of inputs, such a network would have to make large changes to its heuristic model. Thus, improvement in performance is only possible if the network is able to cross certain \textit{semantic barriers}. The notion of semantic barriers motivates a natural choice for transition rates, analogous to thermally activated hopping across an energy barrier. We therefore assume that the rate to transition from model $\mathcal{M}_i$ to $\mathcal{M}_j$ assumes an Arrhenius-like form $R_{ij} = R_{ij}^{0}\rm{exp}\left[-\beta(\mathcal{F}_{j}-\mathcal{F}_i)\right]$. These transition rates in turn imply an Arrhenius-like law for the waiting time $\tau_{ij}$ for the transition from $\mathcal{M}_{i}$ to $\mathcal{M}_j$, which is given by 

\begin{equation} \label{eq:3}
    \tau_{ij} = \tau_{ij}^{0}e^{\beta(\mathcal{F}_{j}-\mathcal{F}_i)}
\end{equation}

The constants $\beta$ and $\tau_{ij}^{0}$ can depend on the dataset size $D$, number of parameters $N$, as well as hyperparameters of the neural network, but the fallibilities themselves do not. In the interest of parsimony, we will assume throughout the rest of the paper, that $\beta$ is independent of $D$ and $N$, and only $\tau_{ij}^{0}$ varies with these quantities. In the next two sections, we will show that even under these relatively mild assumptions, the semantic landscape paradigm can account for several empirical results on scaling and emergence in deep neural networks. 

\section{Explaining emergent phenomena using semantic landscapes}
\subsection{Grokking}
The term `grokking' refers to a phenomenon in which neural networks exhibit delayed generalization, i.e. the validation accuracy remains low during training long after the training accuracy reaches 100\%, before suddenly rising at late times. Grokking was first observed by \cite{power2022grokking} on small algorithmic datasets and has since been observed in a wide range of contexts (\cite{liu2022omnigrok,murty2023grokking}). A number of explanations for grokking have been proposed, based on the behavior of weight norms during training and testing. These include the slingshot mechanism(\cite{thilak2022slingshot}),the LU model (\cite{liu2022omnigrok}), and competition between sparse and dense subnetworks (\cite{merrill2023tale}). Grokking has also been described as a phase transition (\cite{vzunkovivc2022grokking},\cite{liu2022towards}) within the framework of representation learning. \cite{nanda2023progress} have given a fascinating account of the sequence of learning stages that consists of memorization, circuit formation, and cleanup. 

Within the semantic landscape paradigm, whether a given network will exhibit grokking or not depends on the topography of the semantic landscape, which in turn depends on the task that the network is required to perform. If there are relatively few ways of performing the task correctly, and a large number of ways of performing it incorrectly, the semantic landscape will be characterized by one, or a few, very deep minima, surrounded by a sea of relatively shallow ones corresponding to high fallibility, and separated from them by semantic barriers. This type of semantic landscape topography is most likely to result in observable grokking. This is because starting from any of the high fallibility models, the network must be able to locate one of the few solutions that lead to low fallibility (and hence low test loss), which would generally take much longer than it takes to memorize the training data set. This follows from the intuition that memorization is the easiest heuristic model to adopt, as it requires only storage and retrieval of information with no additional processing. For example if a neural network is tasked with classifying dogs and pigeons, memorizing the training images would be easier than developing heuristics that involve formulating meaningful concepts such as "wings" or "legs" by extracting features from those images. This argument provides a simple intuitive explanation for why grokking is more likely to be observed on algorithmic datasets, and is supported by the finding that grokking on realistic datasets, when present, is weaker than on algorithmic ones (\cite{liu2022omnigrok}). Furthermore, since finding solutions that generalize well involves crossing semantic barriers, the semantic landscape paradigm predicts that the test loss during grokking should transiently increase with training steps before decreasing again, which has indeed been observed in studies of grokking (\cite{power2022grokking,nanda2023progress}). The amount by which loss transiently increases contains information about the height of semantic barriers, and is likely to play a crucial role in future work aimed at connecting network weight norm dynamics to the heuristic models discussed here.

An important observation regarding grokking is the presence of a critical training dataset size (\cite{power2022grokking}) below which the network is unable to generalize regardless of how long the training phase lasts. Can the semantic landscape paradigm account for the presence of such a critical datasize? To answer this question, we need to understand how training data affect dynamics on the semantic landscape. To begin with, the presence of training examples is what enables the network to hop between heuristic models in the first place. Supplying more data enables the network to access more models. Thus, the configuration space of heuristic models is a graph whose vertices correspond to heuristic models, and whose connectivity is determined by the training data. If multiple edges are present between two models, we can interpret this as an increased propensity for transitions between those two models. Conversely, if no edges are present between two models, it is impossible to transition between them. In the simplest approximation, we can think of the dataset size $D$ as being proportional to the average degree of the graph of heuristic models. 

\begin{figure}
	\centering
    \includegraphics[width= \textwidth]{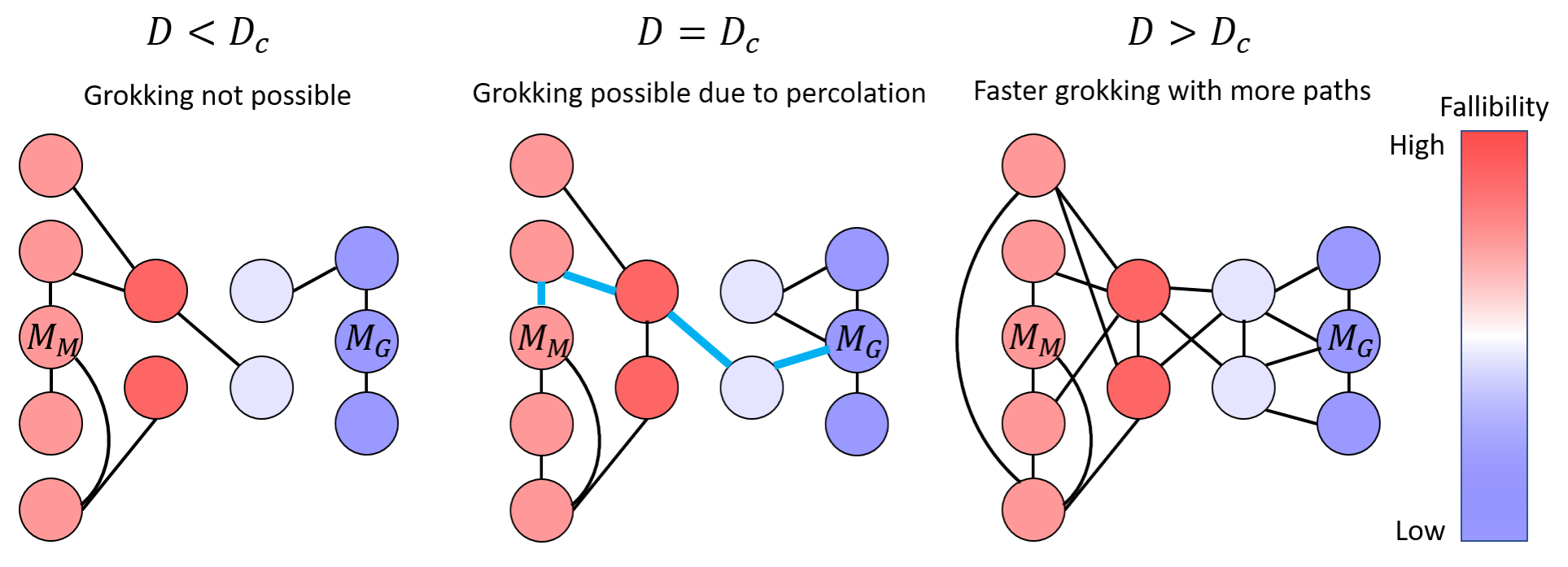}
	\caption{\textbf{Grokking as a percolation phenomenon:} Schematic showing how grokking emerges within the semantic landscape paradigm. Grokking corresponds to a first passage process to reach model heuristic model $\mathcal{M}_G$ starting from $\mathcal{M}_M$ via intermediate semantic barriers corresponding to models with higher $\mathcal{F}$, as indicated by the colorbar. This process is possible only when a continuous path between $\mathcal{M}_M$ and $\mathcal{M}_G$ exists (middle panel, thick blue lines), which happens only if $D\geq D_c$. Increasing $D$ further leads to faster grokking due to increased connectivity of the graph.}
	\label{fig2}
\end{figure}

Let us assume that the network has adopted the heuristic model $\mathcal{M}_M$ that corresponds to memorization of the training dataset. For sufficiently large $D$, there is always a continuous path that connects $\mathcal{M}_M$ to the model corresponding to the generalization solution $\mathcal{M}_G$. However, as we reduce the dataset size, edges between models are progressively removed until we reach a critical dataset size $D_c$ below which the graph becomes disconnected, and it is no longer possible to transition from $\mathcal{M}_M$ to $\mathcal{M}_G$ regardless of how long we run the training (Fig. \ref{fig2}). This implies that the model is incapable of generalization below a critical dataset size, as observed in previous work (\cite{power2022grokking,liu2022towards}). Crucially,this simple argument implies that that the phase transition associated with grokking is a \textit{percolation transition} on the graph of heuristic models. The semantic landscape paradigm also successfully predicts the decrease of grokking time with dataset size above the critical threshold. Since the number of edges of the graph controls the probability of transitioning between nodes, the grokking rate $\lambda_g$ is proportional to the ``excess degree'' of the graph above the percolation threshold, i.e. $\lambda_g \propto (D-D_c)$, and therefore vanishes at $D = D_c$. Correspondingly, the grokking time $t_g \propto 1/(D-D_c)$. Quite remarkably, this scaling argument agrees with theoretical predictions as well as experimental observations in (\cite{liu2022towards}), suggesting a correspondence between the two frameworks. 

\subsection{Emergence of new capabilities with scale}
\begin{figure}
	\centering
    \includegraphics[width= \textwidth]{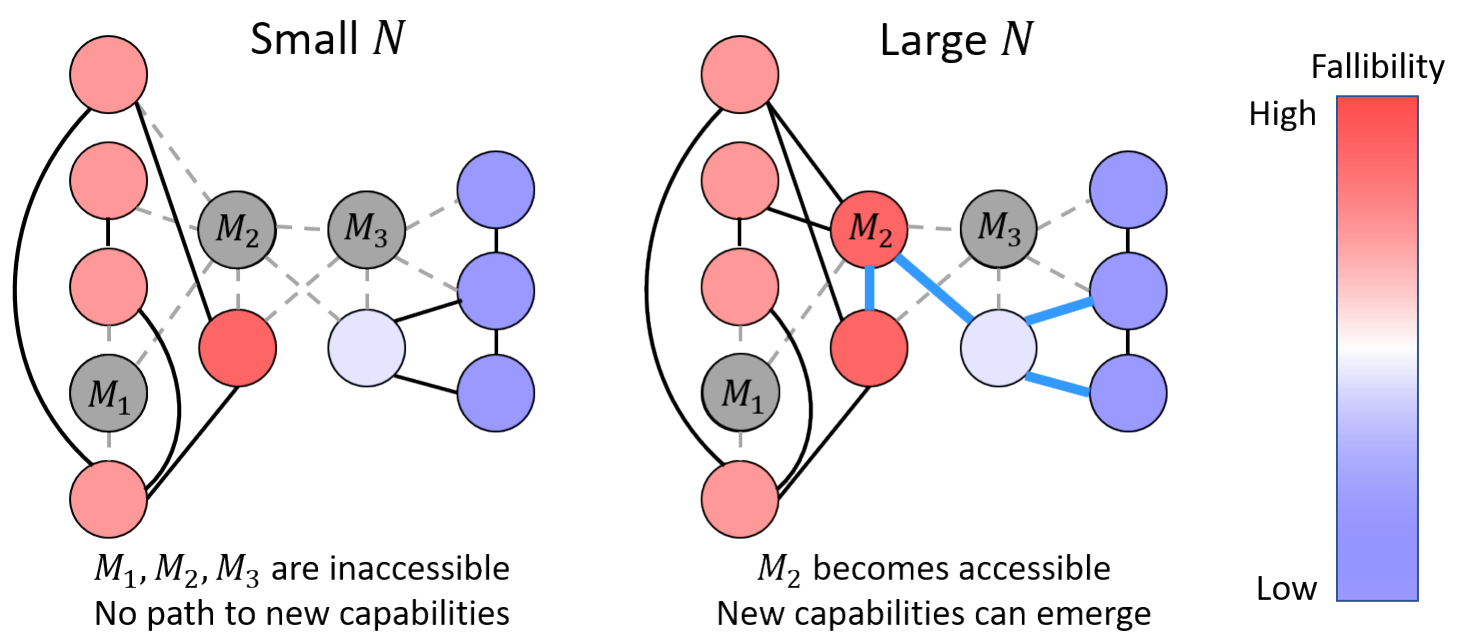}
	\caption{\textbf{Emergence with scale as a percolation phenomenon:} Left panel: For small $N$, the network is incapable of formulating the heuristic models $\mathcal{M}_1$,$\mathcal{M}_2$, and $\mathcal{M}_3$ (grey), and hence, the associated paths (dashed grey lines) are forbidden. Right panel: Increasing $N$ allows the network to formulate model $\mathcal{M}_2$, which opens a path towards low fallibility models, (thick blue lines), and enables the network to learn new capabilities.}
	\label{fig3}
\end{figure}

While grokking refers to the emergence of new capabilities in time, several studies have also reported the sudden emergence of new capabilities with increasing model size, i.e. the number of model parameters $N$ (\cite{brown2020language,rae2021scaling,austin2021program,wei2022emergent,michaud2023quantization}). Like grokking, the unpredictable nature of emergence with scale simultaneously makes it one of the most challenging, as well as the most crucial problems in AI that need to be solved in the near future (\cite{ganguli2022predictability}). Thus far, relatively few studies have offered an explanation for abrupt emergence of new capabilities with scale. It has recently been proposed that grokking and double descent, a particular form of emergence with scale, might be related phenomena (\cite{davies2023unifying}). The most noteworthy effort in this direction is the quantization model of neural scaling (\cite{michaud2023quantization}) which explains emergence in terms of the sequential acquisition of discrete computational capabilities called `quanta'. The semantic landscape paradigm shares some interesting similarities with the quantization model, which we discuss separately in Section $5$. To explain emergence with scale, we need to understand how the semantic landscape is influenced by $N$. Let us assume that the dataset size $D \gg D_c$, so that we are not bottlenecked due to insufficient data. As discussed in Section $2$, increasing $N$ enhances the capacity of the network to formulate new propositions, and hence formulate new heuristic models. Thus, the effect of increasing $N$ is to increase the number of nodes on the graph of heuristic models. If a particular node drops out of the network, all edges connecting it must also drop out, even if the dataset $D$ is infinitely large. Physically, this means that the network is incapable of formulating that particular heuristic model, regardless of how many training examples it sees. Thus, if we consider the configuration space as the graph of \textit{all possible nodes for all $N$}, the nodes corresponding to models that can only be formulated for large $N$ will remain disconnected at small $N$. With this insight in mind, it is easy to see that emergence with scale corresponds to a percolation transition with increasing $N$, such that a continuous path to models of low $\mathcal{F}$ opens up only for sufficiently large $N$ (Fig. \ref{fig3}).  

Percolation can also account for the distinct behaviors of per token loss in large language models (LLMs) observed by \cite{michaud2023quantization}. Even if the network finds a heuristic model of lower $\mathcal{F}$, particularly for complex tasks such as language prediction, it is likely that the improved model does significantly better on some but not all tokens. Some tokens may require the network to perform several successive jumps to increasingly lower $\mathcal{F}$ states for them to be predicted accurately. \cite{michaud2023quantization} refer to these as ``polygenic tokens''. For complex tasks such as language prediction, the semantic landscape is likely to be rugged, consisting of a large number of minima and saddle points, which provides a natural explanation for the presence of polygenic tokens. 

\section{Explaining neural scaling laws using semantic landscapes}
Apart from unpredictable emergent phenomena, neural networks also exhibit predictable power law scaling with increasing $D$, $N$, as well as training steps $S$, as mentioned in Section $1$. The first important experimental observation to note is that neural network performance can be bottlenecked by $D$ as well as $N$, leading to diminishing returns when the other parameter is scaled (\cite{kaplan2020scaling}). This observation is already anticipated by the semantic landscape paradigm, as seen in the previous section. Specifically, if $D$ is a limiting factor, performance plateaus because the network is \textit{unable to find paths} to good heuristic models, even if it has the capacity to formulate them (i.e. $N\to\infty$). If, on the other hand, $N$ is a limiting factor, performance plateaus because the network is \textit{unable to formulate} a good heuristic model, even if the paths leading to it are in principle open (i.e $D\to\infty$). In the following discussion, we will therefore assume that $D$ as well as $N$ are both sufficiently large, such that the graph of heuristic models is always connected. 

\subsection{The semantic trap model}

\begin{figure}
	\centering
    \includegraphics[width= 0.7\textwidth]{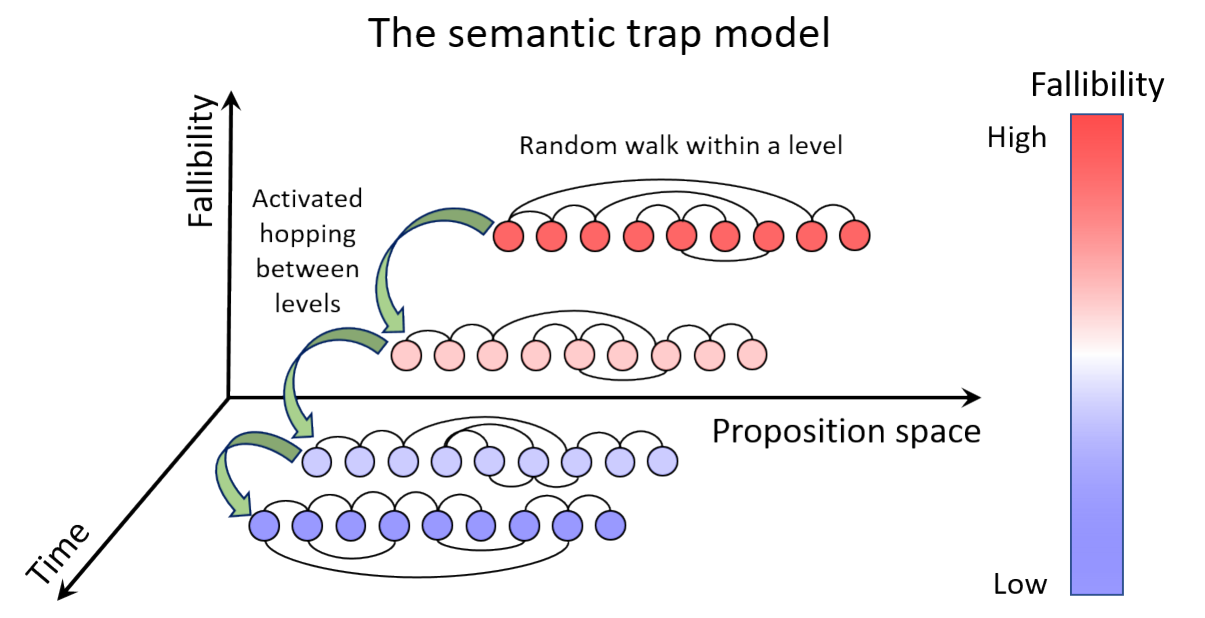}
	\caption{\textbf{Schematic illustration of the semantic trap model.}}
	\label{fig4}
\end{figure}

To derive neural scaling laws within the semantic landscape paradigm, we have to make assumptions about the topography of the semantic landscape. In the case of LLMs, on which most of the work on neural scaling has focused, we expect that the semantic landscape is very rugged, much like the energy landscapes of disordered systems such as structural (\cite{debenedetti2001supercooled}) as well as spin glasses (\cite{mezard1987spin}). At low temperatures, the ruggedness, or the preponderance of metastable energy minima causes the dynamics on such landscapes to proceed sluggishly via activated hops in a non-stationary manner, a process known as aging (\cite{berthier2011theoretical}). One would expect similar dynamics on the semantic landscape of LLMs, as progress towards models with lower $\mathcal{F}$ would presumably be impeded by a large number of semantic barriers. Taking inspiration from Bouchaud's trap model for glassy dynamics (\cite{bouchaud1992weak}), we define the analogous semantic trap model for LLMs. We assume that dynamics on the semantic landscape consists of activated hops across large semantic barriers $\Delta\mathcal{F}_i$ to models with lower $\mathcal{F}$, interspersed with excursions along relatively flat directions that don't affect $\mathcal{F}$ significantly. We further assume that the fallibility $\mathcal{F}$ of the network after $N_b$ barrier crossings is inversely proportional to $N_b$. This model is schematically illustrated in Fig. \ref{fig4}. Analogous to the trap model (\cite{bouchaud1992weak}), we assume that these large semantic barriers have an exponential distribution 

\begin{equation}\label{eq:4}
    \rho(\Delta\mathcal{F}) = \beta\rho_0 \rm{exp}\left(-\beta\mu\Delta\mathcal{F}\right)
\end{equation}

where $\rho_0$,$\beta$, and $\mu$ are constants. In the trap model, $\mu = T/T_g$, where $T$ is the temperature, and $T_g$ is the glass transition temperature. An exponential distribution of barrier heights is known to emerge in a number of disordered systems (\cite{monthus1996models}) and is thought to follow from extreme event statistics (\cite{bouchaud1997universality}). Together with the activated form for dynamics on the semantic landscape (Eqn.\ref{eq:3}), Eqn.\ref{eq:4} implies a waiting time distribution $\psi(\tau)$ that has a power law tail for long waiting times of the form

\begin{equation} \label{eq:5}
    \psi(\tau) = \psi_0\tau_0^{\mu}\tau^{-(1+\mu)}
\end{equation}

where $\psi_0$ is a normalization constant (\cite{bouchaud1992weak}). As each semantic barrier is crossed, the waiting time to cross that barrier is sampled independently, and thus, the time taken to cross $N_b$ barriers is given by $t(N_b) = \sum_i^{N_b}\tau_i$, which results in the scaling $t(N_b) \propto \tau_0 N_b^{1/\mu}$ (\cite{bouchaud1990anomalous}). Assuming that time is measured in the number of training steps $S$, and using $\mathcal{F}\propto 1/N_b$, we obtain the following scaling form of the fallibility with training time

\begin{equation} \label{eq:6}
    \mathcal{F}(S) = \left(\frac{S_0}{S}\right)^{\mu}
\end{equation}

where $S_0$ is a constant. The inverse proportionality between $\mathcal{F}$ and $N_b$ marks a crucial difference between Bouchaud's trap model and the semantic trap model introduced here. Bouchaud's model corresponds to a \textit{renewal process} in which the system loses memory completely following an activated hop, and the energy of the new state is chosen randomly after every hop. By contrast, the semantic trap model predictably reduces its fallibility by an amount proportional to $1/N_b^2$ for the $(N_b+1)^{\rm{th}}$ activated hop.

To obtain the scaling of fallibility with dataset size $D$ and number of parameters $N$, first recall our assumption in Section $2$ that fallibility, and therefore semantic barriers themselves, are independent of $D$ and $N$. The scaling form is therefore determined by the dependence of $\tau_0$ on $D$ and $N$. In the semantic trap model, $1/\tau_0$ is the attempt frequency for activated hops towards lower $\mathcal{F}$, and $\tau_0$ therefore corresponds to the time over which the network explores the sub-graph associated with flat regions of the semantic landscape between hops (horizontal rows of nodes in Fig. \ref{fig4}). From the discussion in Section $3.1$ as well as Fig. \ref{fig2}, $D$ is proportional to the connectivity of the sub-graph. Thus, the rate $\gamma$ at which the sub-graph is traversed increases with $D$. From Section $3.2$ and Fig. \ref{fig3}, we see that $N$ increases the number of heuristic models that the network can formulate. Thus, increasing $N$ increases the number of paths by which the network can find models with lower $\mathcal{F}$. Thus, the typical distance $\ell$ between nodes on the sub-graph that the network has to traverse before it can hop towards a model with lower $\mathcal{F}$ decreases with $N$. For a dynamical process with exponent $\alpha$ on the sub-graph, therefore, we have $\ell(N) \propto (\gamma(D) t)^{\alpha}$. The exponent $\alpha$ characterizes the nature of the process by which the network navigates the graph of heuristic models. Thus, we have $\alpha = 1$ for ballistic dynamics, $\alpha = 0.5$ for diffusive dynamics, $0<\alpha<0.5$ for sub-diffusive dynamics, and $\alpha > 0.5$ for super-diffusive dynamics. Since the network representation takes a finite time to evolve even in the infinite data limit, we assume $\gamma(D) = \gamma_{\infty} D/(D+D_k)$. Similarly, not all models will lead to lower fallibility even for an infinitely large network, and so we have $\ell(N) = \ell_0/N^{1/d_f} + \ell_{\infty}$, where $d_f$ is the fractal dimension of the sub-graph. This scaling form is motivated by the intuition that if $n$ points are randomly distributed within a $d$ dimensional volume, the typical distance between them scales as $n^{-1/d}$. Thus, as $D\to\infty$, we have $\tau_0\propto N^{-1/(\alpha d_f)}$. Furthermore, as $\mathcal{F}\propto 1/N_b$, and $N_b\propto \tau_0^{-\mu}$, we have the following scaling law as $D\to\infty$

\begin{equation} \label{eq:7}
    \mathcal{F}(N)  = \left(\frac{N_0}{N}\right)^{\mu/(d_f\alpha)}
\end{equation}

where $N_0$ is a constant. Finally, as $N\to\infty$, we get $\tau_0 \propto 1/D$, which yields the scaling law

\begin{equation} \label{eq:8}
    \mathcal{F}(D)  = \left(\frac{D_0}{D}\right)^{\mu}
\end{equation}

where $D_0$ is a constant. Eqns. \ref{eq:7} and \ref{eq:8} hold as long as $D_c\ll D \ll D_k$, and $N_c \ll N \ll (\ell_0/\ell_{\infty})^{d_f}$, which is consistent with the range over which typical LLMs operate (\cite{kaplan2020scaling}). Strikingly, from Eqns. \ref{eq:6} and \ref{eq:8}, we see that the semantic trap model predicts that the exponents for dataset scaling and training time scaling should be identical, which is also predicted by the quantization model (\cite{michaud2023quantization}) despite starting from a different set of assumptions, which suggests intriguing connections between the two frameworks. 

\subsection{Interpretation of scaling exponents}
While the semantic trap model is possibly the simplest choice that one can make to describe learning on a rugged semantic landscape, it is certainly possible to construct more sophisticated models. However, despite its simplicity, the semantic trap model can recover a wide range of empirically observed scaling laws (\cite{villalobosscaling,michaud2023quantization}), as the exponents for $D$ and $N$ scaling are independent. In the simplest case, if the landscape is smooth and there are no semantic barriers to cross, the scaling laws are independent of $\mu$. In this case, for ballistic motion $\alpha = 1$ on a regular 2D graph with $d_f=2$, one recovers the variance limited scaling for deep neural networks ($\mathcal{F}(D) \propto 1/D$ and $\mathcal{F}(N) \propto 1/\sqrt{N}$), while $\alpha = 1$, $d_f = 1$ yields the scaling for linear networks ($\mathcal{F}(D) \propto 1/D$ and $\mathcal{F}(N) \propto 1/\sqrt{N}$) (\cite{bahri2021explaining}). We hope that future works will explain this correspondence by systematically coarse-graining from weights to propositions. Comparing $D$ scaling with that observed in large language models (\cite{kaplan2020scaling}), we get $\mu \approx 0.1$. In the trap model (\cite{bouchaud1992weak}), $\mu<1$ corresponds to temperatures lower than the glass transition temperature, where dynamics are extremely sluggish and the system exhibits aging. Curiously, molecular dynamics simulations of glasses have reported that the lowering of potential energy with time is compatible with a power law decay with a small exponent $\approx 0.144$ (\cite{kob1997aging}). Since the scaling exponent for $D$ and time is the same in the semantic trap model, the similarities in the observed scaling in glassy systems and the semantic trap model is quite suggestive, and merits further exploration. 

\section{Discussion}
In the preceding sections, we have introduced and formally defined the notions of heuristic models, fallibility and the semantic landscape. We have demonstrated that a wide range of neural network phenomena including grokking, emergence with scale, and various forms of neural scaling, which have hitherto been explained using disparate theoretical frameworks, find natural and consistent explanations within the semantic landscape paradigm under fairly general assumptions. The main advantage of invoking the semantic landscape paradigm is that neural network phenomena can be recast in terms of well studied problems in statistical physics, such as percolation, spin glasses, and random walks on graphs. By construction, the semantic landscape paradigm interpolates between top-down empirical observations on the high level mechanistic interpretation of neural network function (\cite{nanda2023progress,chughtai2023toy}) obtained via reverse engineering, and bottom-up approaches based on coarse-graining over network parameters \cite{roberts2022principles}. Naturally, at this incipient stage, a discussion of how the semantic landscape paradigm can complement existing approaches, and how it can be developed further, is is order. We therefore conclude this paper with a series of comments that elaborate on these questions and their potential answers.

\subsection{How can we tell whether a phenomenon is emergent or not?}
Since much of this paper was devoted to explaining emergent phenomena, it is worth addressing an interesting contrarian perspective on emergence that was put forward by \cite{schaeffer2023emergent}. The authors show that different loss curve can show either smooth or abrupt evolution with network size depending on the metric used for quantifying loss. Based on this, they argue that emergence may not be a fundamental property of scaling neural network models. We advocate the viewpoint that emergence must necessarily refer to the emergence of new capabilities, and is therefore completely independent of loss metrics. Some metrics could be sensitive to emergence, and others could be insensitive. By contrast, an abrupt change in particular loss metrics doesn't by itself imply that a new capability has emerged. The only real way to determine whether a new capability has emerged is to look under the hood and figure out what has changed in the network's learned representations. Such \textit{mechanistic interpretability} (\cite{nanda2023progress}) analysis might be complicated for complex tasks, not least because the network might be acquiring several new capabilities simultaneously in a way that makes the average loss curve appear smooth (\cite{nanda2022mechanistic}). However, we strongly believe that real progress on understanding emergence can only be made via deciphering the logic developed within the network's internal representations during learning. 

\subsection{Semantic landscapes and the quantization model}
The quantization model (\cite{michaud2023quantization}) is a theoretical framework that is capable of explaining emergence as well as scaling in neural networks. It begins with three hypotheses that can briefly be described as follows: QH1) Neural networks learn a number of discrete computations called `quanta' that are instrumental in reducing loss. QH2) Quanta are learned in order of their usefulness, with more important quanta being learned first. The priority order for learning quanta can therefore be specified in terms of the `Q-sequence'. QH3) The frequency with which quanta are used decays as a power law in the quanta's index in the Q-sequence, which ultimately explains neural scaling laws. The semantic trap model has many interesting similarities, and also some interesting differences with the quantization model. Since heuristic models are algorithms in the network's internal formal language, a heuristic model with lower $\mathcal{F}$ will necessarily contain more quanta. The semantic trap model assumes that $\mathcal{F}$ scales inversely with the number $N_b$ of hops that reduce $\mathcal{F}$. This effectively implies that more useful quanta are learned earlier, as hypothesized in QH2. However, we note that for the scaling relations in Eqn.\ref{eq:6}-\ref{eq:8} to hold, the inverse relationship between $\mathcal{F}$ and $N_b$ only needs to hold on average, and  a weaker version of QH2 that admits occasional disruptions of the Q-sequence should be sufficient. Importantly, the semantic landscape paradigm provides a rigorous definition of the ``usefulness'' of quanta. The ``usefulness'' of a quantum $q$ is exactly equal to the reduction in $\mathcal{F}$ after $q$ has been learned, which is in turn proportional to the number of additional tokens that can be predicted with improved accuracy. This automatically explains why less useful quanta are also ones that are less frequently used. In future work, it would be exciting to ask whether the usefulness of a particular quantum depends on the presence or absence of other quanta, akin to epistatic interactions between mutations during evolution (\cite{wolf2000epistasis}).

Unlike QH3, the semantic trap model does not explicitly assume power laws. Instead, a power law waiting time distribution follows from the assumption of an exponential distribution of semantic barriers. While this choice is still purely phenomenological, it finds precedent in a variety of disordered systems (\cite{monthus1996models}), and could potentially be quite prevalent due to its origin in large deviations theory (\cite{bouchaud1997universality}). Crucially, neural scaling in the semantic landscape paradigm follows from the structure of the semantic landscape, which is at least in principle, formally derivable from network weights. Semantic landscapes therefore serve as a consistent conceptual foundation for making controlled hypotheses in different settings, and can be easily generalized to different network architectures and tasks.  

\subsection{Connecting loss landscapes to semantic landscapes} As noted earlier, the loss landscape picture has proved to be invaluable for our current understanding of neural networks (\cite{mehta2019high,spigler2019jamming,baity2018comparing,bahri2021explaining,roberts2022principles}). From a theoretical perspective, loss landscapes will also prove to be extremely valuable in the construction of semantic landscapes, via systematic coarse-graining of neural weights. In particular, such studies will help assess for which architectures and learning algorithms the assumption of Arrhenius-like transition rates remains valid. However, we stress that several predictions of the semantic landscape paradigm, including the existence of a critical dataset size for grokking, the emergence of new capabilities with scale, as well as power law scaling of loss with network size and dataset size, are determined by the topology of the graph of heuristic models, and do not rely on particular choices of transition rates. The Arrhenius form of transition rates is primarily essential to obtain nontrivial exponents for data scaling. Examining these findings through the lens of the loss landscape will go a considerable way in developing and refining the semantic landscape paradigm in the coming years. 

\subsection{Mechanistic interpretability and the semantic landscape paradigm}
The semantic landscape paradigm provides a consistent and versatile conceptual framework that will prove to be particularly useful for studies that aim to gain mechanistic insights into the learning dynamics of neural networks. Within the last couple of years, a growing number of studies have reverse engineered neural networks to discover circuits that implement a series of computations that enable the network to generalize beyond the training dataset (\cite{nanda2022mechanistic,wang2022interpretability,hernandez2022scaling,chughtai2023toy,lindner2023tracr,conmy2023towards,zhong2023clock}). As this burgeoning field grows and develops further, it would benefit immensely from a framework that keeps track of the training process not in terms of abstract functions of neural weights, but in terms of well-defined and interpretable emergent algorithms. This is exactly what the semantic landscape paradigm offers, by envisioning the training process as a trajectory on the graph of learned algorithms. Furthermore, as emergent algorithms must ultimately derive from weights, mechanistic interpretability studies will be invaluable in developing the semantic landscape paradigm into a more rigorous theory, as they can simultaneously keep track of the microscopic as well as macroscopic aspects of learning. Studies that combine theoretical ideas contained in the semantic landscape paradigm and empirical approaches employed in mechanistic interpretability studies will prove to be instrumental in elucidating the intricacies of neural network learning, thereby paving the way for developing stronger and more efficient AI systems.

\subsection{Is scale everything in AI?}
\cite{ganguli2022predictability} have astutely noted that while the test loss of large generative models improves with increased computing resources and loosely correlates with improved performance on many tasks, specific capabilities cannot be predicted ahead of time. Crucially, while the predictable aspects of performance drive development and deployment of these models, the unpredictable aspects make it difficult to anticipate the consequences of such development and deployment. One could reasonably ask if this is necessarily such a big problem. Indeed, one could argue that this problem might disappear if we simply scale our neural networks up sufficiently, so that whatever capabilities they are currently lacking will be discovered by their scaled up versions. There is no doubt that \textit{quantitatively} speaking, networks with more parameters will acquire more capabilities. However, there is no guarantee that the \textit{qualitative} capabilities that we desire will be found by simply scaling up. In glass physics, the potential energy of glassy states is only slightly higher than that of the global free energy minimum corresponding to the crystalline state (\cite{debenedetti2001supercooled}), and yet glasses differ from crystals significantly in several important properties. Furthermore, it is virtually impossible for glasses to `find' the crystalline state, in the absence of external intervention. If the semantic landscape of large generative models resembles the energy landscape of a glass, they may very well encounter a similar problem. As a corollary, the test loss alone may not be the best indicator of the network's performance. Scaling up the network may not solve this problem because even if the larger network is capable of discovering a solution that the smaller one could not, it may not be able to `find' that solution during its training. Indeed, one may have to tune hyperparameters, engineer prompts, or implement an entirely different, perhaps brain-like (\cite{liu2023seeing}) learning algorithm for it to acquire the desired capability. Ultimately, if we do not understand the algorithms that the network is learning to implement, we would not be able to assess how the network's training process should be modified to achieve desired results. And it is perhaps in solving this critical problem that the confluence of mechanistic interpretability studies and the semantic landscape paradigm will find its finest hour. 

\section*{Acknowledgements}
The author thanks Jeff Gore, Sarah Marzen, Chih-Wei Joshua Liu, Yizhou Liu, Jinyeop Song, Jiliang Hu, and Yu-Chen Chao for illuminating discussions and feedback on the manuscript, as well as Amit Nagarkar and Satyajit Gokhale for a critical reading of the manuscript. The author acknowledges the Gordon and Betty Moore Foundation for support as a Physics of Living Systems Fellow through Grant No. GBMF4513. 

\bibliographystyle{unsrtnat}
\bibliography{references}  %%% Uncomment this line and comment out the ``thebibliography'' section below to use the external .bib file (using bibtex) .

\end{document}